\documentclass[pmlr,twocolumn,10pt]{jmlr} 

\setcitestyle{authoryear,open={(},close={)}}

\usepackage[utf8]{inputenc} 
\usepackage[T1]{fontenc}    
\usepackage{hyperref}       
\usepackage{url}            
\usepackage{microtype}      
\usepackage{nicefrac}       
\usepackage{enumitem}

\usepackage{wrapfig}
\usepackage{float}
\usepackage[skip=0.5pt]{caption} 
\usepackage{chngcntr} 

\usepackage{booktabs}
\firstpageno{1}
\usepackage{titletoc}
\usepackage{titlesec}

\usepackage{subcaption}
\usepackage{makecell}

\newcommand{\stackedmatrix}[2]{
    \begin{bmatrix}
        #1 \\
        #2
    \end{bmatrix}
}
\newcommand{\stackednormal}[6]{
    \mathcal{N} \left(
    \left[\begin{array}{c}
        #1 \\
        #2
    \end{array}\right],
    \left[\begin{array}{cc}
        #3 & #4 \\
        #5 & #6
    \end{array}\right]
    \right)
}
\usepackage{siunitx}

\jmlrvolume{225}
\jmlryear{2023}
\jmlrsubmitted{LEAVE UNSET}
\jmlrpublished{LEAVE UNSET}
\jmlrworkshop{Machine Learning for Health (ML4H) 2023} 

\title[A Probabilistic Method to Predict Classifier Accuracy]{A Probabilistic Method to Predict Classifier Accuracy\\on Larger Datasets given Small Pilot Data}

\author{
  \Name{Ethan Harvey}$^1$\Email{ethan.harvey@tufts.edu}\\
  \Name{Wansu Chen}$^2$\Email{wansu.chen@kp.org}\\
  \Name{David M. Kent}$^3$\Email{david.kent@tuftsmedicine.org}\\
  \Name{Michael C. Hughes}$^1$\Email{michael.hughes@tufts.edu}\\
  \addr$^1$Department of Computer Science, Tufts University, Medford, MA, USA\\
  \addr$^2$Department of Research and Evaluation, Kaiser Permanente Southern California, Pasadena, CA, USA\\
  \addr$^3$Predictive Analytics and Comparative Effectiveness Center, Tufts Medical Center, Boston, MA, USA
}

\begin{document}
\setlength{\abovedisplayskip}{2pt plus 3pt}
\setlength{\belowdisplayskip}{2pt plus 3pt}

\maketitle

\begin{abstract}
Practitioners building classifiers often start with a smaller pilot dataset and plan to grow to larger data in the near future.
Such projects need a toolkit for extrapolating how much classifier accuracy may improve from a 2x, 10x, or 50x increase in data size.
While existing work has focused on finding a single ``best-fit'' curve using various functional forms like power laws, 
we argue that modeling and assessing the \emph{uncertainty} of predictions is critical yet has seen less attention.
In this paper, we propose a Gaussian process model to obtain probabilistic extrapolations of accuracy or similar performance metrics as dataset size increases.
We evaluate our approach in terms of error, likelihood, and coverage across six datasets.
Though we focus on medical tasks and image modalities, our open source approach\footnote{We open source our code at \href{https://github.com/tufts-ml/extrapolating-classifier-accuracy-to-larger-datasets}{\texttt{https://github.com/}}\newline\href{https://github.com/tufts-ml/extrapolating-classifier-accuracy-to-larger-datasets}{\texttt{tufts-ml/extrapolating-classifier-accuracy-to-}}\newline\href{https://github.com/tufts-ml/extrapolating-classifier-accuracy-to-larger-datasets}{\texttt{larger-datasets}}} generalizes to any kind of classifier.
\end{abstract}
\begin{keywords}
Learning curve; Gaussian process
\end{keywords}

\section{Introduction}
\label{sec:intro}

\begin{figure}[t!]
\floatconts
  {fig:motivation}
  {\caption{Example learning curves for predicting infiltration from chest x-rays assessed using area under the receiver operating characteristic (AUROC). Left: Single ``best-fit'' using power law \citep{rosenfeld2020constructive}. Right: Our probabilistic Gaussian process with a power law mean function and 95\% confidence interval for uncertainty.}}
  {\includegraphics[width=1\linewidth]{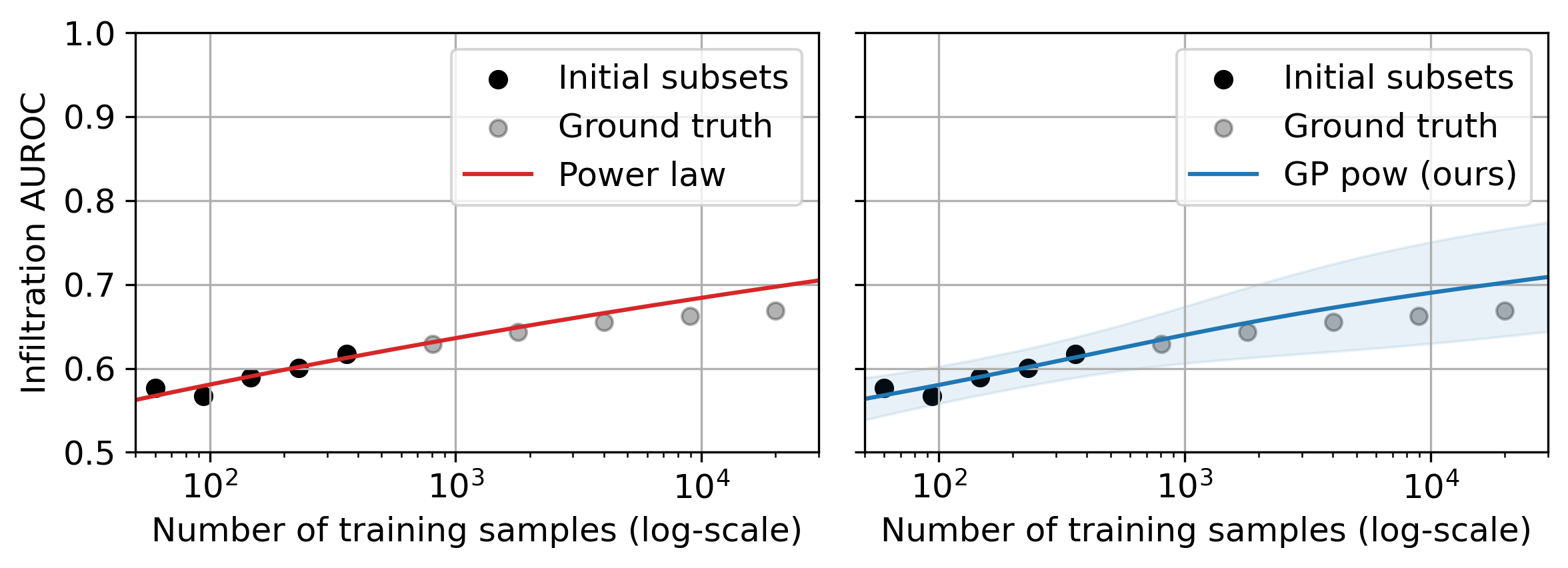}}
\end{figure}

Consider the development of a medical image classifier for a new diagnostic task.
In this and other applications of supervised machine learning, the biggest key to success is often the size of the available labeled training set.
When a large dataset of labeled images is not available, research projects often have a common trajectory: (1) gather a small ``pilot'' dataset of images and corresponding class labels, (2) train classifiers using this available data, and then (3) plan to collect an even larger dataset to further improve performance.
When gathering more labeled data is expensive, practitioners face a key decision in step~3: 
\emph{given that the classifier's accuracy is y\% at the current size $x$, how much better might the model do at $2x$, $10x$, or $50x$ images?}

Despite decades of research, practitioners lack standardized tools to help answer this question of how to extrapolate classifier performance to larger datasets. 
We argue that tools that help \emph{manage uncertainty} are especially needed.
As illustrated in the left panel of Fig.~\ref{fig:motivation}, recent approaches have focused almost entirely on estimating one single ``best-fit'' curve, using power laws \citep{hestness2017deep, rosenfeld2020constructive}, piecewise power laws~\citep{jain2023meta}, or other functional forms~\citep{mahmood2022much}.
In practice, however, no single curve fit to a limited set of size-accuracy pairs can extrapolate perfectly.
Probabilistic methods that can model a \emph{range} of plausible curves, as illustrated in Fig.~\ref{fig:motivation} (right), are thus a more natural solution.
Surprisingly, however, most existing methods focus on deterministic rather than probabilistic modeling (see Tab.~\ref{tab:related-work}). 
The few methods that do explain how to provide probabilistic intervals for their predictions lack careful evaluation of their ability to quantify uncertainty.

In this work, we provide a portable modeling toolkit to help practitioners extrapolate classifier performance \emph{probabilistically}. 
We pursue a Bayesian approach via an underlying Gaussian process (GP) model \citep{rasmussen2006gaussian}.
The mean function of our GP can match common curve forms like power law or arctan, but is adjusted to encode desirable inductive biases (e.g., more data implies better performance) as well as common sense (e.g., accuracy or AUROC can never exceed 100\%).
We further use prior distributions over model hyperparameters to encode domain-specific knowledge (e.g., where might accuracy saturate for this task given enough data).
The whole pipeline remains learnable from a handful of size-accuracy exemplars gathered on a small pilot dataset.

To summarize, our contributions are: (1) a reusable GP-based accuracy probabilistic extrapolator (APEx-GP) that can match existing curve-fitting approaches in terms of error while providing additional uncertainty estimates, and (2) a careful assessment of our proposed probabilistic extrapolations compared to ground truth on larger datasets across six medical classification tasks involving both 2D and 3D images across diverse modalities (x-ray, ultrasound, and CT) with various sample sizes.
While we focus on image analysis tasks here, nothing about our methodology is specialized to images; our pipeline could be repurposed for tabular data, genomics, text, time series, or even heterogeneous data from many domains.
Ultimately, we hope our methods help research teams and sponsoring funding agencies assess data adequacy for proposed research studies.

\section{Background}
\label{sec:background}
We wish to build a classifier for a task of interest using a bespoke dataset.
We assume the largest available labeled dataset $\mathcal{D}$ for our task has \emph{limited size}, which we operationalize as roughly 500-20000 total images with corresponding labels.
We partition $\mathcal{D}$ into non-overlapping training and test sets.
Given a chosen classifier and a specific data partition with $x$ training images, we fit the classifier (including hyperparameter search) then evaluate on that partition's test set, obtaining a performance value $y \in [0.0, 1.0]$. 
We'll assume throughout that higher $y$ implies a better model; we will informally refer to $y$  as ``accuracy'' for convenience, though $y$ could measure any common classifier metric like AUROC or balanced accuracy where 1.0 means ``perfect''.

To estimate how $y$ changes with dataset size using available data, we construct a handful of nested \emph{subsets}, following~\citet{mahmood2022much, jain2023meta}.
First, pick $R$ desired train-set sizes $\{x_r\}_{r=1}^R$ in increasing order.
Next, stochastically sample training sets $\mathcal{S}_1 \subset \mathcal{S}_2 \subset \ldots \subset \mathcal{S}_R$, such that $|\mathcal{S}_r| = x_r$ for each index $r$.
Also sample a non-overlapping test set.
Finally, fit the classifier to each train set $\mathcal{S}_r$ and record the performance on the test set as $y_r$.
Averaging over $y_r$ from multiple random partitions of $\mathcal{D}$ into train and test sets can obtain smoother estimates of heldout performance at each data size $x_r$.

\textbf{Problem statement.}
We now present two possible formulations of our extrapolation problem.

\emph{(i) Point estimate extrapolation.}
Given a small dataset of $R$ size-accuracy pairs $\{x_r, y_r\}_{r=1}^R$, fit a function $f_{\theta}(x)$ so that we can extrapolate classifier accuracy on larger datasets with size $x_* > x_R$.

\emph{(ii) Probabilistic extrapolation.}
Given a small dataset of $R$ size-accuracy pairs $\{x_r, y_r\}_{r=1}^R$, fit a probability density function $p_{\theta}( Y | x)$ treating accuracy $Y$ as the random variable so that we can extrapolate classifier accuracy on larger datasets with appropriate uncertainty.

\setlength{\tabcolsep}{4pt}
\begin{table*}[t!]
\floatconts
  {tab:related-work}
  {\caption{Related work focused on predicting model performance.}}
  {\small\begin{tabular}{llll}
  \toprule
  \bfseries Related work & \bfseries \makecell[l]{Models uncertainty?} & \bfseries \makecell[l]{Evaluates\\uncertainty?} & \bfseries \makecell[l]{Validated on\\medical data?} \\
  \midrule
  Power law {\scriptsize \citep{rosenfeld2020constructive}} & No & N/A & No\\
  \makecell[tl]{Arctan + other functions {\scriptsize \citep{mahmood2022much}}} & No & N/A & No\\
  \makecell[tl]{Learn-Optimize-Collect {\scriptsize \citep{mahmood2022optimizing}}}
    & \makecell[tl]{Post-hoc PDF over $x$ not $y$} 
    & No & No\\
  \makecell[tl]{Piecewise power law {\scriptsize \citep{jain2023meta}}}
    & \makecell[tl]{Yes, but PDF $p(y|x)$ \\ via asymptotic formulas} & No & No\\
  APEx-GP (ours) & Yes, via direct model & Yes & Yes\\
  \bottomrule
  \end{tabular}}
\end{table*}
\setlength{\tabcolsep}{6pt}

\textbf{Evaluation metrics.}
Given a heldout set of size-accuracy pairs $x_*, y_*$, we consider several evaluation metrics.
The first applies to both deterministic (i) and probabilistic (ii) methods. The latter two are only for probabilistic methods.

\begin{itemize}
  \item Error (i or ii). For each heldout pair $x_*, y_*$, assess the error between $y_*$ and $f_{\theta}(x_*)$, via \emph{root mean squared error} or \emph{mean absolute error}.
  
  \item Quantized likelihood (type ii only). For each pair $x_*, y_*$, we compute the probability mass assigned to a narrow interval around the true observation $p( Y \in (y_* - \Delta, y_* + \Delta) | x_*)$, with $\Delta=0.01$. Assessing an interval ensures this metric is always between 0 and 100\% (higher is better).
  
  \item Coverage (type ii only) \citep[p.~93]{dodge2003oxford}. 
  Here we assume that for each $x_*$ we observe many replicates of $y_*$ (via re-sampling train-test splits). From the model's probability density function (PDF) $p( Y | x_*)$ we obtain an interval $y_a, y_b$ corresponding to the P\% high-density interval. We then measure the fraction of times the measured replicates of $y_*$ falls in that interval: this empirical fraction should match $P$\% if the model is well-calibrated.
\end{itemize}

\section{Related Work}
\label{sec:related}

\paragraph{Data scaling.}

Prior works have empirically validated that generalization in deep learning scales with dataset size according to a power law function in both computer vision \citep{sun2017revisiting, bahri2021explaining, hoiem2021learning, zhai2022scaling} and natural language processing \citep{hestness2017deep, kaplan2020scaling}.
\citet{sun2017revisiting} find that image classification accuracy increases logarithmically based on the training dataset size.
\citet{hestness2017deep} show test loss decreases according to a power law as training dataset size increases in machine translation, language modeling, image processing, and speech recognition.
\citet{zhai2022scaling} scale vision transformer (ViT) \citep{dosovitskiy2021vit} models and data, both up and down, and find the power law characterizes the relationships between error rate, data, and compute.

\paragraph{Predicting model performance.}

Other works look at predicting model performance at larger dataset sizes \citep{cortes1993learning, frey1999modeling, johnson2018predicting, rosenfeld2020constructive, jain2023meta} and estimating data requirements given performance targets \citep{mahmood2022much, mahmood2022optimizing} (see Tab.~\ref{tab:related-work}).
\citet{rosenfeld2020constructive} develop a model for predicting performance given a specified model. They find that errors are larger when extrapolating from smaller dataset sizes. 
\citet{jain2023meta} propose a piecewise power law that models performance as a quadratic curve in the few-shot setting and a linear curve in the high-shot setting.
They estimate confidence intervals using a formula from ~\citet{gavin2019levenberg} inspired by estimators of covariance matrices for parameters fit by maximum likelihood~\citep{murphy2022GaussApproxMLE}. However, such estimators are only justified \emph{asymptotically} as sample size increases; use when estimating from only a few data points seems questionable.

\citet{mahmood2022much} consider a broad class of computer vision tasks and systematically investigate a family of functions that generalize the power law function to allow for better estimation of data requirements. They focus on estimating target data requirements given an approximate relationship between data size and model performance; such as a power law function.
\citet{mahmood2022optimizing} propose a new paradigm for modeling the data collection workflow as a formal optimal data collection problem that allows designers to specify performance targets, collections costs, a time horizon, and penalties for failing to meet the targets.
They estimate the distribution of $x$ that achieves the target performance $y$ by bootstrapping size-accuracy pairs, estimating the dataset size, and fitting a density estimation model; in contrast, we directly model uncertainty in $y$.

\paragraph{Sample size estimation.}
Loosely related to our work are traditional sample size estimation calulations.
\citet{riley2020calculating} provide practical guidance for calculating the sample size required for the development of clinical prediction models.
These include calculations that might identify datasets that are too small (for example, if overall outcome risk cannot be estimated precisely).

\section{Probabilistic Model}
\label{sec:methods}

We now develop our approach to modeling a probability density function $p(y|x)$ that can estimate the distribution in accuracy $y$ at any specific training set size $x$.

\subsection{GP extrapolation model}
\label{sec:GP_overview}

For each possible dataset size $x$, we imagine there is an unobservable random variable $f$ representing \emph{true} classifier performance, as well as an observable random variable $y$ representing \emph{realized} classifier performance on a finite test set.
To achieve a flexible model for function $f(x)$, we turn to a Gaussian process prior with mean function $m$ and covariance function $k$.
Given true accuracy $f$, we then model each observable accuracy $y$ as a perturbation of true accuracy $f$ by independant and identically distributed (IID) Gaussian noise with scalar variance $\tau^2$.
When we condition on a finite set of data size inputs $\mathbf{x} = \{x_r\}_{r=1}^R$ of interest, our model's joint distribution $p(\mathbf{y}, \mathbf{f} | \mathbf{x})$ factorizes as
\begin{align}
  p(\mathbf{f} \mid \mathbf{x}) &= \mathcal{N}(\mathbf{f} \mid m(\mathbf{x}), k(\mathbf{x}, \mathbf{x})) \\
  p(\mathbf{y} | \mathbf{f}, \mathbf{x}) &= \textstyle \prod_{r=1}^R \mathcal{N}(y_r | f_r, \tau^2) \nonumber 
\end{align}
where $\mathbf{f}, \mathbf{y}$, and $m(\mathbf{x})$ are each $R$-dimensional vectors whose entry at index $r$ corresponds to the provided data size input $x_r$.
Similarly, $k(\mathbf{x}, \mathbf{x})$ is an $R \times R$ covariance matrix, with entry $s,t$ equal to $k(x_s, x_t)$.

Below, we provide recommended options for both mean $m$ and covariance function $k$.
We pay particular attention to the mean, offering two concrete choices, a power law and an arctan, inspired by the best performing methods from prior work on point estimation of ``best-fit'' curves for size-accuracy extrapolation.
Fig.~\ref{fig:meanfunc_vs_x} shows both possible mean functions across a range of parameters $\theta$ while saturating at a maximum accuracy of $1-\varepsilon$.
Other uses of GPs in practice often assume a constant mean of zero; we select forms that deliberately allow accuracy to grow as more data is added.

\begin{figure}[t!]
\floatconts
  {fig:meanfunc_vs_x}
  {\caption{Example varying parameters for the power law (left) and arctan function (right) with $\varepsilon = 0.05$.}}
  {\includegraphics[width=1\linewidth]{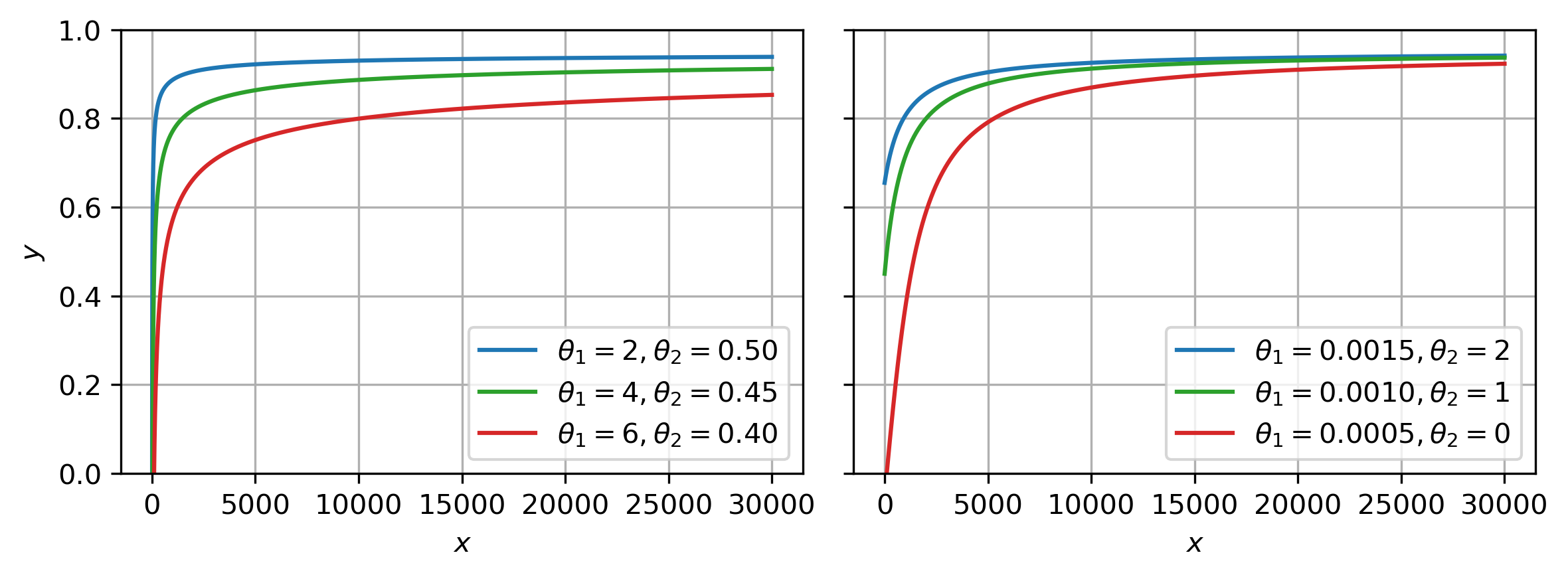}}
\end{figure}

\paragraph{Power law mean.} Our power law function is
\begin{align}
m^{\text{pow}}(x) = (1-\varepsilon) - \theta_1 x^{\theta_2},
\end{align}
with two trainable parameters $\theta_1 \geq 0$ and $\theta_2 \in [-1, 0]$.
Similar power law forms have been previously recommended~\citep{cortes1993learning, frey1999modeling, johnson2018predicting, rosenfeld2020constructive}.
Our version guarantees that $\lim_{x \rightarrow \infty} m(x) = 1 - \varepsilon$.

\paragraph{Arctan mean.} Our arctan mean function is
\begin{align}
m^{\text{arc}}(x) =& \frac{2}{\pi} \textrm{arctan}\left( \theta_1 \frac{\pi}{2} x + \theta_2 \right) - \varepsilon
\end{align}
with two trainable parameters: $\theta_1 \ge 0$ and $\theta_2 \ge 0$. 
Similar forms were recommended by recent work~\citep{mahmood2022much}.
Our modified version guarantees that $\lim_{x \rightarrow \infty} m(x) = 1 - \varepsilon$.

\paragraph{Encoding saturation limits via $\varepsilon$.}
Our definition of both power law and arctan means above deliberately includes an $\varepsilon$ term to allow domain experts to define how the function should saturate as the training set size grows $x \rightarrow \infty$.
Setting $\varepsilon = 0$ allows a ``perfect classifier'' with $m(x) = 1$, while setting $\varepsilon = 0.05$ reflects a lower ceiling that may be more appropriate. 
Even with an infinite training set, we might not be able to build a perfect classifier (e.g., due to image noise, label noise, insufficiency of images alone for the diagnostic task, and interrater reliability issues).

\paragraph{Covariance function.}
The classic radial basis function (RBF) kernel~\citep{rasmussen2006gaussian} applied to \emph{logarithms} of input sizes defines our covariance:
\begin{align}
k(x, x') = \sigma^2\exp\left(-\frac{ (\log(x) - \log(x')) ^2}{2\lambda^2}\right)
\end{align}
where both the output-scale $\sigma > 0$ and length-scale $\lambda > 0$ are trainable parameters.
This log-RBF form implies that $f(x)$ and $f(x')$ values have high covariance at similar sizes $x$ and $x'$ , while at very different sizes (numerator $\gg 2\lambda^2$) the $f$ values have near-zero covariance.
We select the log-RBF because it tends to produce \emph{smooth} functions $f(x)$, and we expect the idealized classifier accuracy as a function of data size to be smooth~\citep{mahmood2022much, mahmood2022optimizing}.
We expect selecting other stationary kernels like Matern or avoiding the $\log$ would have relatively minor impact on overall model quality, as long as output scale and length scale are trainable.

\subsection{Prior control of extrapolation uncertainty}
\label{sec:methods_prior}
Our GP model has two kinds of parameters.
Let $\mathbf{\eta} = \{\tau, \sigma, \lambda, \varepsilon \}$ denote the parameters that control \emph{uncertainty} (in the likelihood or the GP covariance function) or \emph{asymptotic behavior} (e.g., $\varepsilon$ sets the saturation value of $m(x)$).
We argue that domain knowledge can and should guide learning of $\mathbf{\eta}$.
In this section, we develop prior distributions for each parameter in $\mathbf{\eta}$.
In contrast, parameters $\mathbf{\theta} = \{\theta_1, \theta_2 \}$ define the shape of the mean function and thus are more straightforward to estimate even given a small training set of $R$ size-accuracy pairs. We do not define priors for $\theta$.

\textbf{Likelihood scale $\tau$.}
The scalar $\tau > 0$ represents the standard deviation of each realized accuracy $y$ given ideal accuracy $f$.
We expect \emph{a priori} that realized accuracy $y$ does not vary too much around $f$; any deviation of more than 0.03 seems undesirable.
We operationalize our desiderata (details in App.~\ref{apd:second}) with a truncated normal prior \citep{fisher1931truncated}.

\textbf{Kernel output-scale $\sigma$.}
Scalar $\sigma > 0$ controls the magnitude of variance for random variable $f$.
Along with $\tau$, it also controls the variance of accuracy $y$ if $f$ is marginalized away (as we do in extrapolation).
Given only a small dataset of size-accuracy pairs, we \emph{a priori} should have considerable uncertainty about realized accuracy $y(x_*)$ at sizes $x_*$ much larger than seen in the pilot set. 
This matches past empirical evidence ~\citep{rosenfeld2020constructive, mahmood2022much, jain2023meta}.
We operationalize the prior as a truncated normal and do a numerical grid search for mean and variance hyperparameters such that a desired wide 20-80 percentile range is satisfied (details in App.~\ref{apd:second}).

\textbf{Kernel length-scale $\lambda$.}
Scalar $\lambda > 0$ controls the rate at which the correlation between $f(x)$ and $f(x')$ decreases as the distance between $\log(x)$ and $\log(x')$ increases.
A reasonable \emph{a priori} belief is that strong correlations may only exist when the distance between $x$ and $x'$ is less than $1.5x$; at larger distances we should not expect strong correlation as the dataset would have nearly doubled in size.
We chose a truncated normal prior that which gives roughly the desired behavior (details in App.~\ref{apd:second}).

\textbf{Saturation limit $\varepsilon$.}
Scalar $\varepsilon \ge 0$ represents one minus the maximum accuracy we expect as dataset size gets asymptotically large, using domain knowledge.
Given plausible lower and upper bounds from domain experts (see App.~\ref{apd:first} and \ref{apd:second}), we form a uniform prior over $\varepsilon$.

\subsection{Fitting to data via MAP estimation}
\label{sec:methods_fit}
Given a training set of $R$ size-accuracy pairs, represented by sizes $\mathbf{x}$ and accuracies $\mathbf{y}$, fitting the model means estimating the parameters $\theta$ and $\eta$, defined early in Sec.~\ref{sec:methods_prior}.
We can take advantage of our model's conjugacy to integrate away our latent variable $\mathbf{f}$. This leaves the marginal likelihood of the observable training set as:
\begin{align}
    p_{\theta, \eta}(\mathbf{y} | \mathbf{x}) &= \int_{\mathbf{f}} p_{\eta}(\mathbf{y} | \mathbf{f}, \mathbf{X}) p_{\theta, \eta}(\mathbf{f} | \mathbf{X}) d\mathbf{f}.
\end{align}
We then optimize the following maximum a-posteriori (MAP) objective to obtain point estimates of $\theta
$ and $\eta$:
\begin{align}
\label{eq:map_objective}
\hat{\theta}, \hat{\eta} = \arg\!\max_{\theta, \eta} ~\log p_{\theta,\eta}( \mathbf{y} | \mathbf{x} ) + \log p(\eta).
\end{align}
The objective works in log space for numerical stability, where the log of the marginal likelihood has the following closed-form
\begin{align}
    \log p(\mathbf{y}|\mathbf{x}) =& - \frac{R}{2}\log2\pi - \frac{1}{2}\log| \mathbf{K} + \tau^2 I_R| \\
    & - \frac{1}{2}(\mathbf{y} - \mathbf{m})^T( \mathbf{K} + \tau^2 I_R)^{-1} (\mathbf{y} - \mathbf{m}). \nonumber
\end{align}
Here, $\mathbf{K}$ is a $R \times R$ matrix defined as $ k(\mathbf{x}, \mathbf{x})$ and depends on $\sigma, \lambda \in \eta$.
Vector $\mathbf{m}$ is defined as $ m(\mathbf{x})$ and depends on $\theta$ and $\varepsilon$. We suppressed the explicit dependence on $\theta,\eta$ in this notation for simplicity.

\subsection{Extrapolation via the posterior predictive}
\label{sec:methods_predict}

Given a fit model via estimates $\hat{\theta}, \hat{\eta}$, our target use for our model is to extrapolate probabilistically. Conditioning on a pilot training set of $R$ size-accuracy pairs $\mathbf{x}, \mathbf{y}$, we wish to estimate the posterior over accuracies $y_*$ at a set of $Q$ larger dataset sizes $x_*$.
Again using well-known properties of GPs, 
specifically the joint-to-conditional transformation of Gaussian variables (details in App.~\ref{apd:third}), we can directly compute the PDF of the posterior predictive
\begin{align}
\label{eq:extrapolation}
    p( \mathbf{y}_* &| \mathbf{x}_*, \mathbf{y}, \mathbf{x} ) =\ \mathcal{N}( \mathbf{y}_* | \mathbf{\mu}, \mathbf{\Sigma}), \\ 
    &\mathbf{\mu} = \mathbf{m}_* + \mathbf{K}^T_* (\mathbf{K} + \tau^2 I_R)^{-1} \left(\mathbf{y} - \mathbf{m} \right) \nonumber \\
    &\mathbf{\Sigma} = \mathbf{K}_{**} + \tau^2 I_{Q} - \mathbf{K}^T_* (\mathbf{K} + \tau^2 I_R)^{-1} \mathbf{K}_*. \nonumber
\end{align}
Here $K_*$ is a $R \times Q$ matrix and $K_{**}$ is a $Q \times Q$ matrix, where each entry is a call to covariance function $k$ with appropriate inputs from the train or test sets. All calls to $m$ or $k$ use estimated parameters $\hat{\theta}, \hat{\eta}$.

\paragraph{Constraining $y$ to [0,1].} A careful reader will note that in our application the ``accuracy'' $y$ must be a positive real confined to the interval [0.0, 1.0]. In contrast, throughout our modeling derivation we allow $y$ broader support over the whole real line. We chose this broader support for the computational convenience it provides at training time: our training objective in Eq.~\eqref{eq:map_objective} and our posterior predictive in Eq.~\eqref{eq:extrapolation} are both computable in closed-form. 
To avoid extraneous predictions, for any scalar invocation of Eq.~\eqref{eq:extrapolation} after forming the univariate posterior prediction, we truncate the predictive density to the unit interval [0.0, 1.0].
This support broadening has statistical justification \citep{wojnowicz2023approximate}.

\section{Experimental Procedures}
\label{sec:experiments}

We now describe how we gather the size-accuracy pairs used to train and evaluate models from various 2D and 3D medical imaging datasets.
We then outline procedures for assessing predictions for both short-range (up to $2x$ the train set size) and long-range (up to $50x$) settings.

\subsection{Datasets and Classifier Procedures}

Our chosen 2D datasets all use 224x224 resolution and span x-ray and ultrasound modalities, including ChestX-ray14 \citep{wang2017chestx}, \citet{kermany2018labeled}'s Chest X-Ray Pneumonia dataset, the Breast Ultrasound Image dataset (BUSI) \citep{al2020dataset}, and the Tufts Medical Echocardiogram Dataset (TMED-2) \citep{huang2021new, huangTMED2Dataset2022}.
We also study two 3D datasets of head CT scans: the Open Access Series of Imaging Studies (OASIS-3) dataset \citep{lamontagne2019oasis} and a proprietary pilot neuroimaging dataset.
See App.~\ref{apd:fourth} for detailed dataset descriptions and preprocessing steps.

From each dataset, we form one or more separate binary classification tasks, using only labels with sufficient data (at least 10\% prevalence).
When raw labels are multiclass, we transform to several one-vs-rest binary tasks. This choice allows using the same pipeline and same evaluation metrics in all analyses, substantially simplifying the work and presentation.

All datasets contain fully deidentified images gathered during routine care. Only the last one is not public.
The use of these deidentified images for research has been approved by our Institutional Review Board (Tufts Health Science IRB \#11953).

\paragraph{Classifiers.}
For all tasks, we fine-tune the classification head of a ViT pretrained on ImageNet \citep{deng2009imagenet}.
For 2D tasks, the pretrained ViT processes each 224x224 image and produces an image-specific embedding $h_i \in \mathbb{R}^D$, where $D=768$. We then model the binary label of interest $C_i$ given $h_i$ as
\begin{align}
C_i | h_i \sim \text{Bern}(C_i \mid \sigma(w^T h_i)),
\end{align}
\noindent where $w \in \mathbb{R}^D$ are learnable weights and $\sigma$ is the sigmoid function. 

For 3D tasks, we feed each 224x224 2D slice into the pretrained ViT, apply a linear per-slice classifier, and then aggregate via mean pooling to produce a scan-level prediction.
Let $h_{i,n} \in \mathbb{R}^D$ denote ViT embedding of the $n$-th slice (out of $N$) of the $i$-th image, where $D=768$.
We model binary label $C_i$ given all embeddings $\mathbf{h}_{i,1:N}$ as
\begin{align}
C_i | \mathbf{h}_{i,1:N} \sim \textrm{Bern}(C_i \mid \textstyle \frac{1}{N} \sum_{n=1}^N \sigma(w^T h_n)),
\end{align}
where again $w \in \mathbb{R}^D$ are learnable weights and $\sigma$ is the sigmoid function.
While other flexible 3D architectures are possible, we chose this path for simplicity.

\paragraph{Training details.}
For all tasks, we fit weights via MAP estimation, maximizing the above Bernoulli likelihoods plus a Gaussian prior on $w$ (also known as weight decay).
We fit using L-BFGS for 2D and SGD with momentum for 3D.
For grayscale images, we replicate to 3 channels to feed into the 3-channel pretrained ViT.
For 3D images, we reduce computational costs by subsampling at most 50 slices.

\begin{table*}[t!]
\floatconts
  {tab:likelihood}
  {\caption{Quantized likelihood evaluations at heldout $x,y$ pairs for short-range and long-range extrapolations, with $\Delta=0.01$ (see Sec.~\ref{sec:background}). Standard deviations generated from 500 bootstrapping rounds to select data-split seeds to average across. We bold values with non-overlapping intervals. As a baseline, we form a uniform distribution from the minimum accuracy observed in training up to the task-specific maximum accuracy (Sec.~\ref{apd:first}).
  }}
  {\small\begin{tabular}{llr@{.}lr@{.}lr@{.}lr@{.}lr@{.}l}
  \toprule
  & & \multicolumn{2}{c}{} & \multicolumn{4}{c}{Short-range extrapolation} & \multicolumn{4}{c}{Long-range extrapolation}\\
  \bfseries Dataset & \bfseries Label & \multicolumn{2}{c}{\bfseries Baseline} & \multicolumn{2}{c}{\bfseries GP pow} & \multicolumn{2}{c}{\bfseries GP arc} & \multicolumn{2}{c}{\bfseries GP pow} & \multicolumn{2}{c}{\bfseries GP arc}\\
  \midrule
  ChestX-ray14 & Atelectasis & $6$ & $1\pm0.0\%$ & $45$ & $2\pm4.5\%$ & $44$ & $3\pm4.8\%$ & $\mathbf{29}$ & $\mathbf{2}\boldsymbol{\pm}\mathbf{2.6}\boldsymbol{\%}$ & $22$ & $1\pm2.6\%$ \\
  & Effusion & $6$ & $2\pm0.0\%$ & $37$ & $7\pm4.8\%$ & $38$ & $3\pm4.4\%$ & $15$ & $3\pm1.9\%$ & $15$ & $2\pm2.0\%$ \\
  & Infiltration & $4$ & $6\pm0.0\%$ & $\mathbf{44}$ & $\mathbf{7}\boldsymbol{\pm}\mathbf{3.8}\boldsymbol{\%}$ & $24$ & $0\pm5.4\%$ & $\mathbf{25}$ & $\mathbf{2}\boldsymbol{\pm}\mathbf{2.4}\boldsymbol{\%}$ & $1$ & $1\pm1.4\%$ \\
  Chest X-Ray & Bacterial & $11$ & $3\pm0.0\%$ & $42$ & $3\pm8.0\%$ & $42$ & $5\pm7.7\%$ & $38$ & $2\pm10.5\%$ & $43$ & $5\pm8.6\%$ \\  
  & Viral & $6$ & $4\pm0.0\%$ & $39$ & $8\pm6.2\%$ & $38$ & $9\pm6.6\%$ & $12$ & $3\pm5.2\%$ & $\mathbf{24}$ & $\mathbf{6}\boldsymbol{\pm}\mathbf{6.9}\boldsymbol{\%}$ \\
  BUSI & Normal & $20$ & $3\pm0.0\%$ & $48$ & $8\pm9.4\%$ & $48$ & $8\pm9.1\%$ & \multicolumn{2}{c}{---} & \multicolumn{2}{c}{---} \\
  & Benign & $8$ & $5\pm0.0\%$ & $27$ & $9\pm10.8\%$ & $28$ & $7\pm11.2\%$ & \multicolumn{2}{c}{---} & \multicolumn{2}{c}{---} \\
  & Malignant & $15$ & $1\pm0.0\%$ & $27$ & $0\pm13.3\%$ & $27$ & $5\pm12.5\%$ & \multicolumn{2}{c}{---} & \multicolumn{2}{c}{---} \\
  TMED-2 & PLAX & $20$ & $8\pm0.0\%$ & $65$ & $6\pm1.5\%$ & $64$ & $5\pm1.9\%$ & $\mathbf{64}$ & $\mathbf{8}\boldsymbol{\pm}\mathbf{1.0}\boldsymbol{\%}$ & $39$ & $0\pm1.6\%$ \\
  & PSAX & $9$ & $6\pm0.0\%$ & $62$ & $6\pm1.5\%$ & $62$ & $6\pm1.5\%$ & $\mathbf{63}$ & $\mathbf{2}\boldsymbol{\pm}\mathbf{1.1}\boldsymbol{\%}$ & $58$ & $4\pm1.4\%$ \\
  & A4C & $14$ & $3\pm0.0\%$ & $\mathbf{62}$ & $\mathbf{8}\boldsymbol{\pm}\mathbf{2.4}\boldsymbol{\%}$ & $56$ & $1\pm3.4\%$ & $\mathbf{58}$ & $\mathbf{2}\boldsymbol{\pm}\mathbf{3.2}\boldsymbol{\%}$ & $24$ & $5\pm3.8\%$ \\
  & A2C & $8$ & $5\pm0.0\%$ & $18$ & $0\pm2.5\%$ & $\mathbf{61}$ & $\mathbf{3}\boldsymbol{\pm}\mathbf{0.7}\boldsymbol{\%}$ & $24$ & $9\pm2.8\%$ & $20$ & $8\pm3.2\%$ \\
  OASIS-3 & Alzheimer’s & $4$ & $6\pm0.0\%$ & $22$ & $5\pm12.5\%$ & $23$ & $6\pm12.3\%$ & \multicolumn{2}{c}{---} & \multicolumn{2}{c}{---} \\
  Pilot neuro- & WMD & $6$ & $0\pm0.0\%$ & $29$ & $3\pm14.0\%$ & $27$ & $8\pm14.6\%$ & \multicolumn{2}{c}{---} & \multicolumn{2}{c}{---} \\
  imaging dataset & CBI & $5$ & $9\pm0.0\%$ & $29$ & $6\pm14.8\%$ & $27$ & $4\pm14.4\%$ & \multicolumn{2}{c}{---} & \multicolumn{2}{c}{---} \\
  \bottomrule
  \end{tabular}}
\end{table*}

\subsection{Experimental protocol}

For each dataset, we randomly assign images at an 8:1:1 ratio into training, validation, and testing sets, ensuring each patient’s data belongs to exactly one split and stratifying by class to ensure comparable class frequencies.
We repeat this process with three data-split random seeds; each seed selects a different train, validation, and test partition. 

Given a split's particular training set, we form $R=5$ log-spaced subsets with 60 to 360 training samples: $\{60, 94, 147, 230, 360\}$.
We select these subset sizes because small pilot datasets used for demonstrating feasibility typically have only a few hundred samples. Log-spacing captures macro trends rather than micro fluctuations.
We then evaluate approaches for \emph{short-range extrapolation} on five log-spaced subsets between 360 and 720 training samples $\{414, 475, 546, 627, 720\}$, and \emph{long-range extrapolation} on five log-spaced subsets between 360 and 20000 training samples $\{804, 1796, 4010, 8955, 20000\}$.
For both short-range and long-range, we omit any values beyond total available training set size.

When fitting each model on each train-set size, we tune hyperparameters including weight initialization seed, weight decay, and number of epochs (to approximate early stopping, see App.~\ref{apd:fifth}), selecting the configuration that maximizes validation performance.

At each train-set size $x$, we record as ``accuracy'' $y$ the average AUROC. We average across 3 data-split seeds for 2D (15 for 3D) to mitigate high-variance estimates from small test sets.
We can then finally fit extrapolation models and assess error (RMSE), quantized likelihood, and coverage (see Sec. \ref{sec:background}) using these $x,y$ pairs.

\paragraph{Dense Coverage evaluations.}
For the two largest 2D datasets, ChestX-ray14 and TMED-2, we evaluate coverage on $100$ replicates of the above protocol using \emph{long-range coverage} train-set sizes of $\{5\text{k}, 10\text{k}, 20\text{k}\}$ training samples. 
Each accuracy value $y_*$ still represents the mean test performance across three distinct data-split seeds.
Having 100 replicates at each train-set size $x_*$ allows better estimation of the coverage percentage $P\%$.
We include single point coverage evaluations for each dataset in App.~\ref{apd:tenth}.

\begin{figure*}[htbp!]
\floatconts
  {fig:main-figure}
  {\caption{Long-range extrapolation results for atelectasis, effusion, and infiltration from the ChestX-ray14 dataset; bacterial and viral pneumonia from the Chest X-Ray dataset; and PSAX, A4C, and A2C from the TMED-2 dataset.}}
  {\includegraphics[width=1\linewidth]{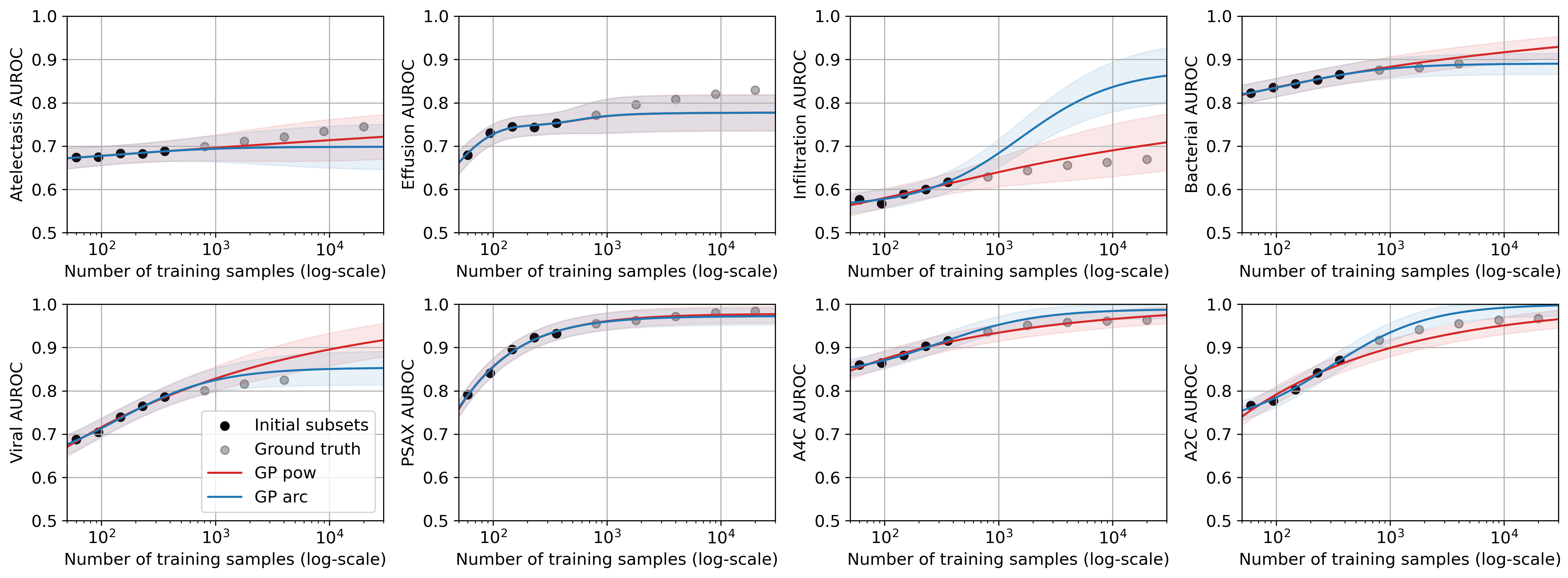}}
\end{figure*}

\setlength{\tabcolsep}{2pt}
\begin{table*}[t!]
\floatconts
  {tab:dense-coverage-95}
  {\caption{Coverage rates for long-range extrapolations, using a target interval of 95\% from our GP model and 100 replicates. Standard deviations generated from 500 bootstrapping rounds to select data-split seeds to average across.}}
  {\footnotesize\begin{tabular}{llr@{.}lr@{.}lr@{.}lr@{.}lr@{.}lr@{.}lr@{.}l}
  \toprule
  & & \multicolumn{4}{c}{5k training samples} & \multicolumn{4}{c}{10k training samples} & \multicolumn{6}{c}{20k training samples} \\
  \bfseries Dataset & \bfseries Label & \multicolumn{2}{c}{\bfseries GP pow} & \multicolumn{2}{c}{\bfseries GP arc} & \multicolumn{2}{c}{\bfseries GP pow} & \multicolumn{2}{c}{\bfseries GP arc} & \multicolumn{2}{c}{\bfseries \makecell{GP pow\\w/o priors}} & \multicolumn{2}{c}{\bfseries GP pow} & \multicolumn{2}{c}{\bfseries GP arc} \\
  \midrule
  ChestX-ray14 & Atelectasis & $100$ & $0\pm0.0\%$ & $100$ & $0\pm0.0\%$ & $100$ & $0\pm0.0\%$ & $100$ & $0\pm0.1\%$ & $55$ & $4\pm2.8\%$ & $100$ & $0\pm0.0\%$ & $99$ & $7\pm0.5\%$ \\
  & Effusion & $100$ & $0\pm0.2\%$ & $99$ & $9\pm0.3\%$ & $94$ & $4\pm1.9\%$ & $91$ & $0\pm2.1\%$ & $0$ & $0\pm0.0\%$ & $48$ & $9\pm3.0\%$ & $39$ & $6\pm2.8\%$ \\
  & Infiltration & $100$ & $0\pm0.0\%$ & $0$ & $0\pm0.0\%$ & $100$ & $0\pm0.0\%$ & $0$ & $0\pm0.0\%$ & $14$ & $5\pm2.6\%$ & $100$ & $0\pm0.0\%$ & $0$ & $0\pm0.0\%$ \\
  TMED-2 & PLAX & $100$ & $0\pm0.0\%$ & $98$ & $5\pm1.1\%$ & $100$ & $0\pm0.0\%$ & $83$ & $2\pm2.6\%$ & $100$ & $0\pm0.0\%$ & $100$ & $0\pm0.0\%$ & $45$ & $0\pm2.9\%$ \\
  & PSAX & $100$ & $0\pm0.0\%$ & $100$ & $0\pm0.0\%$ & $100$ & $0\pm0.0\%$ & $100$ & $0\pm0.0\%$ & $100$ & $0\pm0.0\%$ & $100$ & $0\pm0.0\%$ & $100$ & $0\pm0.0\%$ \\
  & A4C & $100$ & $0\pm0.0\%$ & $27$ & $3\pm2.8\%$ & $100$ & $0\pm0.0\%$ & $20$ & $5\pm2.7\%$ & $100$ & $0\pm0.1\%$ & $100$ & $0\pm0.1\%$ & $20$ & $7\pm2.7\%$ \\
  & A2C & $98$ & $5\pm1.1\%$ & $0$ & $2\pm0.4\%$ & $100$ & $0\pm0.0\%$ & $0$ & $1\pm0.3\%$ & $100$ & $0\pm0.0\%$ & $100$ & $0\pm0.0\%$ & $0$ & $2\pm0.4\%$ \\
  \bottomrule
  \end{tabular}}
\end{table*}
\setlength{\tabcolsep}{6pt}

\section{Results}
\label{sec:results}

Using the procedures described above, we performed extensive experiments designed to answer several key research questions.
First, ``which mean function performs best?''
Second, ``in terms of predictive error, is there a substantial difference between our probablistic approach and previous deterministic approaches?'' 
Finally, ``is the coverage obtained by our probabilistic approach compelling?''

Our major findings are highlighted below.

\paragraph{For short-range extrapolation (up to 2x train size), both mean functions seem competitive.}
Tab.~\ref{tab:likelihood} reports quantized likelihoods (higher is better).
Looking at short-range results, both mean functions perform similarly.
The difference between power law and arctan in 12 of 15 tasks is less than 2\%.

\paragraph{For long-range extrapolation (from 2x-50x train size), the power law mean function seems best.}
Looking at the long-range results in Tab.~\ref{tab:likelihood},
power law wins clearly in 5 of 6 cases and essentially ties in the other cases.
In the other case where arctan wins, power law still clearly outperformas a simpler baseline (arctan can be worse than this baseline sometimes).
Power law's superiority in long-range settings is also supported by coverage results in Tab.~\ref{tab:dense-coverage-95}, and RMSE results in App.~\ref{apd:sixth}.
Unlike the power law, the arctan function seems to produce learning curves that asymptote quickly. This results in minimal change in predicted performance after 5000 samples and in some cases overestimates of performance (see curve for infiltration on ChestX-ray14 in Fig.~\ref{fig:main-figure}).

\paragraph{Error from our GP models is competitive with deterministic models.}
By design, our GP models use mean functions shown by past work to be effective deterministic predictors.
We therefore intend that in terms of pointwise error metrics like root mean squared error, our GP approach is indistinguishable from the non-probabilistic ``best-fit'' curve approach of past works.
RMSE results in App.~\ref{apd:sixth} suggest that on 13 of 15 short-range tasks and 7 of 9 long-range tasks, our GP power law model is either clearly better or within 0.04 RMSE of deterministic power law.

\paragraph{Our chosen priors improve long-range coverage over no priors.}
In Tab.~\ref{tab:dense-coverage-95}, we include coverage at 20k training samples from our GP power law model without priors. 
When performance is low and there is room for variation in performance as dataset size grows, coverage with priors is significantly better than without.
However, when performance is high at small dataset sizes coverage with priors is just as good as without (see coverage for TMED-2 dataset).

\paragraph{Our GP power law model tends to have decent coverage, but over-estimates the intervals a bit.}
Looking at coverage in Tab.~\ref{tab:dense-coverage-95}, our GP power law model consistently achieves 100\% coverage, over-estimating a well calibrated interval.
Although wider intervals are preferred to narrow ones that miss the truth, we emphasize that our goal in Tab.~\ref{tab:dense-coverage-95} is to achieve \textit{95}\% coverage.

\section{Discussion and Conclusion}
\label{sec:conclusion}

We introduced a portable GP-based probabilistic modeling pipeline for classifier performance extrapolation that can match existing curve-fitting approaches in terms of error while providing additional uncertainty estimates.
We compared our probabilistic extrapolations to ground truth on 2-50x larger datasets across six medical classification tasks involving both 2D and 3D images across diverse modalities (x-ray, ultrasound, and CT).
We recommend the power law mean function based off its superior long-range error, quantized likelihood, and coverage.

\textbf{Limitations.}
We acknowledge that APEx-GP is not universally preferred over previous deterministic alternatives; on some tasks (effusion on ChestX-ray14) our error was worse than the power law baseline.
More work is needed to understand if our model would be effective beyond the data modalities, train set sizes, classifier architectures, and AUROC metric used here.
However, we designed our approach to be effective out-of-the-box for other ``accuracy''-like metrics that satisfy two properties: higher is better and 1.0 is a ``perfect'' score.

\textbf{Outlook.}
We hope our approach provides a useful tool for practitioners in medical imaging and beyond to manage uncertainty when assessing data adequacy.

\acks{This work was supported by NIH grant R01-NS102233, recently renewed as 2 RF1 NS102233-05, as well as a grant from the Alzheimer’s Drug Discovery Foundation.}

\bibliography{harvey23}

\appendix

\startcontents[sections]

\counterwithin{table}{section}
\setcounter{table}{0}
\counterwithin{figure}{section}
\setcounter{figure}{0}

\makeatletter 
\let\c@table\c@figure
\let\c@lstlisting\c@figure
\let\c@algorithm\c@figure
\makeatother

\section*{Appendix Contents}
\printcontents[sections]{l}{1}{\setcounter{tocdepth}{2}}

\section{GP Model Details}
\subsection{Plausible Upper and Lower Bounds}
\label{apd:first}
Given plausible lower and upper bounds from domain experts, we form a uniform prior over $\varepsilon$.
We can elicit a plausible upper bound for $\varepsilon$ as $1-y'$, where $y'$ is the maximum accuracy observed in the pilot set.
We can elicit a plausible lower bound by talking with task experts, which we define as $\varepsilon_{\text{min}}$.
Based off plausible upper bounds from domain experts, we use $\varepsilon_{\text{min}} = 0.05$ for 3D datasets of head CT scans. For all other datasets we use $\varepsilon_{\text{min}} = 0.0$.

\subsection{Chosen Priors}
\label{apd:second}
However, the variation we model for $y(x_*)$ should be \emph{at most} the width $W$ of interval $[y', 1.0]$, where $y'$ is the largest accuracy observed in pilot set.
We typically expect less deviation than $W$: we suggest that the marginal of $y(x_*)$, whose variance is approximately $s^2 = \tau^2 + \sigma^2$ if $x_* \gg x_R$, should have a 3-standard-deviation window $w$ whose 20-80 percentile range is between $\frac{W}{2}$ and $\frac{3W}{4}$.
We therefore seek a prior on $\sigma$ such that if we draw many samples of $\sigma$ as well as many samples of $\tau$ from its prior above, the implied window has the desired properties:
\begin{align}
\textsc{Prctile}(w, 20) &\approx \textstyle \frac{W}{2},
\textsc{Prctile}(w, 80) \approx \textstyle \frac{3W}{4}.
\\ \notag 
\text{where~} w \gets 6 s, s &\gets \sqrt{\tau^2 + \sigma^2}, \tau \sim p(\tau), \sigma \sim p(\sigma)
\end{align}
We operationalize the prior as a truncated normal $p(\sigma) = \mathcal{N}_{[0, \infty)}( \mu_{\sigma}, \nu_{\sigma})$, and do a numerical grid search for hyperparameters $\mu_{\sigma}$ and $\nu_{\sigma}$ such that our desired 20-80 range is satisfied.
Figure \hyperref[fig:margin]{\ref{fig:margin}} shows the implied distribution on $w$ (the two-sided 3-std.-dev. window for $y_*$) given our chosen prior $p(\sigma)$.

\begin{figure}[htbp!]
\floatconts
  {fig:margin}
  {\caption{PDF for the margin of $95\%$ of the posterior distribution based off the prior on $\tau$, a clinically informed maximum performance of $0.95$, a test performance of $0.7$ for the small initial dataset, and the $20^{\text{th}}$ and $80^{\text{th}}$ percentile of the margin of the majority of the posterior density to be a fourth and a half of the maximum performance minus the performance of the small initial dataset. The $20^{\text{th}}$ and $80^{\text{th}}$ percentile of the distribution are $0.0625$ and $0.125$, as desired. The histogram was generated with $10$ million samples from the prior on $\tau$ and $\sigma$.}}
  {\includegraphics[width=0.75\linewidth]{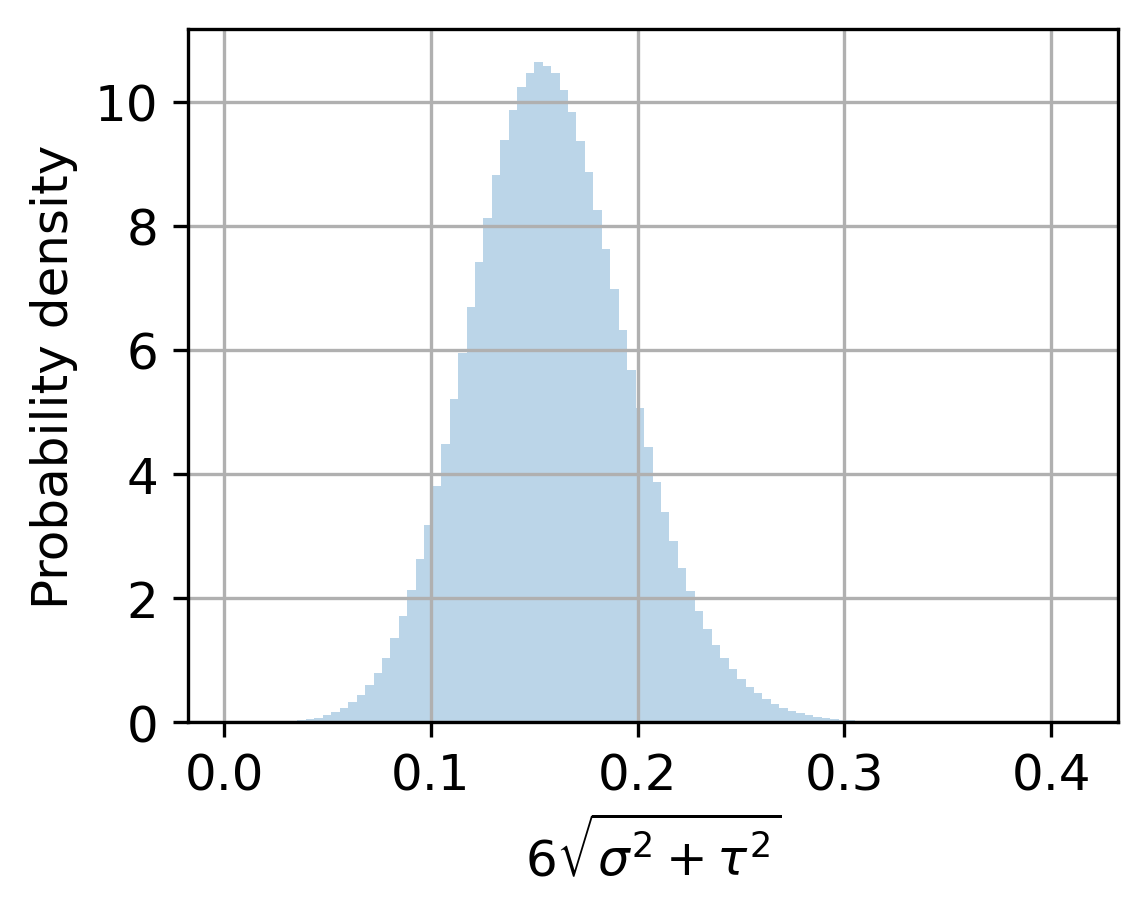}}
\end{figure}

\paragraph{Kernel length-scale $\lambda$.}
Scalar $\lambda > 0$ controls the rate at which the correlation between $f(x)$ and $f(x')$ decreases as the distance between $\log(x)$ and $\log(x')$ increases.
Setting $x' = rx$ for $r \geq 1$, the distance becomes $\log(x') - \log(x) = \log(r)$.
A reasonable \emph{a priori} belief is that strong correlations may only exist when $r$ is less than 1.5; at larger distances we should not expect strong correlation as the dataset would have nearly doubled in size.
We thus seek a prior $p(\lambda)$ whose 10th percentile is around $\lambda=0.13$ (implying a low covariance $k(r) = 0.01\sigma^2$) and 90th percentile  around $\lambda = 2.86$ (implying a high covariance of $k(r) = 0.99 \sigma^2$). 
These desired $\lambda$ values were obtained by solving for $\lambda > 0$ in the relation
\begin{align}
 k(r) = \sigma^2 \exp \Big( -\frac{1}{2\lambda^2} \log(r) ^2 \Big),
\end{align}
with $r = 1.5$.
Concretely, we chose a truncated normal prior $p(\lambda) = \mathcal{N}_{[0,\infty)}(-1.23, 2.14^2)$, which gives roughly the desired behavior.

\paragraph{Saturation limit $\varepsilon$.}
Scalar $\varepsilon \ge 0$ represents one minus the maximum accuracy we expect as dataset size gets asymptotically large, using domain knowledge.
We can elicit a plausible upper bound for $\varepsilon$ as $1-y'$, where $y'$ is the maximum accuracy observed in the pilot set.
We can elicit a plausible lower bound by talking with task experts, which we define as $\varepsilon_{\text{min}}$.
For example, we may not expect to beat accuracy of 0.95 on some task due to inherent interrater reliability issues.
Given these two bounds, we form a uniform prior over $\varepsilon$.

\begin{figure}[htbp!]
\floatconts
  {fig:epsilon}
  {\caption{Example PDF for $\varepsilon$ prior with $\varepsilon_{\text{min}} = 0.05$ and $y' = 0.7$.}}
  {\includegraphics[width=0.75\linewidth]{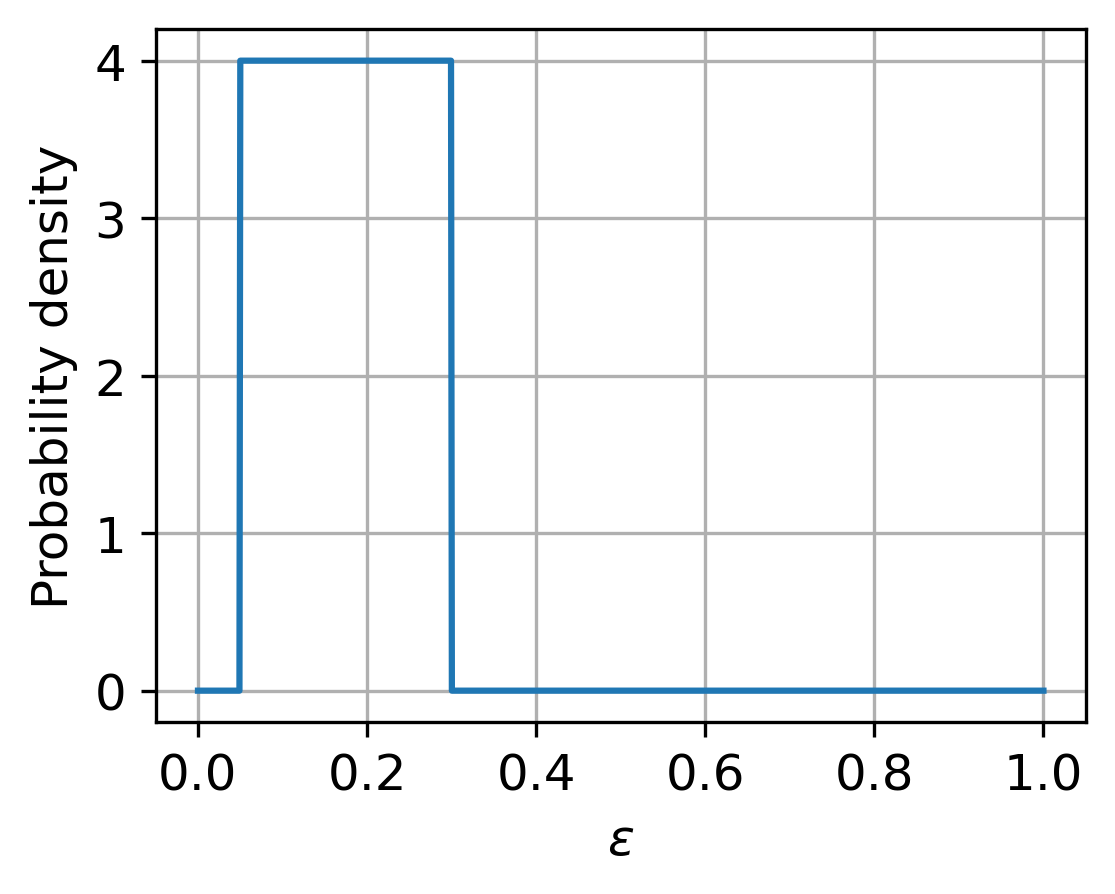}}
\end{figure}

\subsection{Conditional Gaussian Math}
\label{apd:third}

Given a marginal Gaussian distribution for $\mathbf{f}$ and a conditional Gaussian distribution for $\mathbf{y}$ given $\mathbf{f}$

\begin{align*}
\stackedmatrix{\mathbf{f}}{\mathbf{f}_*} &\sim \stackednormal{\mathbf{m}}{\mathbf{m}_*}{\mathbf{K}}{\mathbf{K}_*}{\mathbf{K}^T_*}{\mathbf{K}_{**}} \\
\stackedmatrix{\mathbf{y}}{\mathbf{y}_*} | \stackedmatrix{\mathbf{f}}{\mathbf{f}_*} &\sim \stackednormal{\mathbf{f}}{\mathbf{f}_*}{\tau^2 I_\mathcal{R}}{\mathbf{0}}{\mathbf{0}}{\tau^2 I_\mathcal{Q}}
\end{align*}

the marginal distribution of $\mathbf{y}$ is given by

\begin{align*}
\stackedmatrix{\mathbf{y}}{\mathbf{y}_*} \sim \stackednormal{\mathbf{m}}{\mathbf{m}_*}{\mathbf{K}+\tau^2 I_\mathcal{R}}{\mathbf{K}_*}{\mathbf{K}^T_*}{\mathbf{K}_{**}+\tau^2 I_\mathcal{Q}}.
\end{align*}

\section{Dataset Details}
\label{apd:fourth}

\textbf{ChestX-ray14} is an open access dataset comprised of 112,120 de-identified frontal view X-ray images of 30,805 unique patients with fourteen text-mined disease image labels.
For preprocessing we resize the images to $224 \times 224$ pixels, rescale pixel values to $[0.0, 1.0]$, and normalize pixel values with the source mean and standard deviation.

\textbf{Chest X-Ray} is an open access dataset comprised of 5,856 de-identified pediatric chest X-Ray images.
For preprocessing we center-crop the images using a window size equal to the length of the shorter edge, resize them to $224 \times 224$ pixels, rescale pixel values to $[0.0, 1.0]$, and normalize pixel values with the source mean and standard deviation.
We evaluate our Gaussian process' extrapolation performance for the classification of bacterial and viral pneumonia.

\textbf{BUSI} is an open access dataset comprised of 780 de-identified breast ultrasound images from 600 female patients with an average image size of $500 \times 500$ pixels.
For preprocessing we center-crop the images using a window size equal to the length of the shorter edge, resize them to $224 \times 224$ pixels, rescale pixel values to $[0.0, 1.0]$, and normalize pixel values with the source mean and standard deviation.
The images are categorized into three classes, which are normal, benign, and malignant.

\textbf{OASIS-3} is a open access project aimed at making neuroimaging datasets freely available to the scientific community.
The dataset includes 895 de-identified CT scans from 610 patients where the patient has a diagnosis at least 80 days before or up to 365 days after the CT scan was taken.
We use these diagnoses for binary classification.
Each 3D scan contains a variable number (74-148) of $512 \times 512$ transverse slices.
The images are provided in Hounsfield Units (HU). For preprocessing we skull strip images (only including -100 to 300 HU) \citep{muschelli2019recommendations}, resize each 3D scan to $N \times 224 \times 224$ voxels (where $N$ is the number of transverse slices), rescale pixel values to $[0.0, 1.0]$, and normalize pixel values with the source mean and standard deviation.
We do not correct gantry-tilt since each image's degree of gantry-tilt is not include in the header.

\textbf{TMED-2} is a clinically-motivated benchmark dataset for computer vision and machine learning from limited labeled data.
The dataset includes 24964 de-identified echocardiogram images with view labels from 1280 patients. We use these view labels for binary classification.
For preprocessing we resize the images to $224 \times 224$ pixels, rescale pixel values to $[0.0, 1.0]$, and normalize pixel values with the source mean and standard deviation.

\textbf{Pilot neuroimaging dataset} is a sample of de-identified CT scans from 600 patients. 500 scans were randomly sampled from the cohort of patients 50+ years of age who reveived MRI in 2009-2019 and 100 were randomly sampled from the cohort members who had covert brain infarction (CBI) and/or white matter disease (WMD).
This yielded a total sample that included 142 CBI cases and 156 WMD cases.
The dataset includes scans from multiple planes for each patient in the Digital Imaging and Communications in Medicine (DICOM) CT format.
To simplify the input of our model, we use the largest scan from the axial plane for each patient.
Each 3D scan contains a variable number (23-373) of $512 \times 512$ transverse slices.
For preprocessing we correct gantry-tilt, convert images into HU using each image's rescale slope and intercept, skull strip images, resize each 3D scan to $N \times 224 \times 224$ voxels (where $N$ is the number of transverse slices), rescale pixel values to $[0.0, 1.0]$, and normalize pixel values with the source mean and standard deviation.

\section{Classifier Details}
\label{apd:fifth}

\paragraph{2D datasets.}
We tune hyperparameters including weight initialization seed, weight decay, and number of epochs to maximize validation AUROC.
We select the weight initialization seed from 5 different seeds and weight decay from 11 logarithmically spaced values between $1e^{5}$ to $1e^{-5}$.

\paragraph{3D datasets.} 
We tune hyperparameters including learning rate, weight initialization seed, weight decay, and number of epochs to maximize validation AUROC.
We select the learning rate from 0.05 and 0.01, weight initialization seed from 5 different seeds, and weight decay from 6 logarithmically spaced values between $1e^{0}$ to $1e^{-5}$, as well as without weight decay.

\onecolumn
\newpage

\section{Results: RMSE}
\label{apd:sixth}

\begin{table*}[h!]
\floatconts
  {tab:rmse-short}
  {\caption{AUROC RMSE for short-range extrapolations.}}
  {\small\begin{tabular}{llS[table-format=1.3]S[table-format=1.3]S[table-format=1.3]S[table-format=1.3]}
  \toprule
  \bfseries Dataset & \bfseries Label & \multicolumn{1}{c}{\bfseries Power law} & \multicolumn{1}{c}{\bfseries GP pow (ours)} & \multicolumn{1}{c}{\bfseries Arctan} & \multicolumn{1}{c}{\bfseries GP arc (ours)}\\
  \midrule
  ChestX-ray14 & Atelectasis & 0.344 & 0.329 & 0.397 & 0.396 \\
  & Effusion & 0.673 & 0.859 & 0.671 & 0.811 \\
  & Infiltration & 0.320 & 0.433 & 2.264 & 2.192 \\
  Chest X-Ray & Bacterial & 0.212 & 0.225 & 0.164 & 0.169 \\
  & Viral & 1.012 & 1.046 & 1.095 & 1.095 \\
  BUSI & Normal & 0.705 & 0.705 & 0.705 & 0.705 \\
  & Benign & 1.539 & 1.544 & 1.543 & 1.547 \\
  & Malignant & 1.003 & 1.003 & 1.003 & 1.003 \\
  TMED-2 & PLAX & 0.124 & 0.126 & 0.261 & 0.261 \\
  & PSAX & 0.447 & 0.447 & 0.450 & 0.451 \\
  & A4C & 0.408 & 0.408 & 0.721 & 0.721 \\
  & A2C & 2.177 & 2.174 & 0.081 & 0.082 \\
  OASIS-3 & Alzheimer’s & 1.561 & 1.563 & 1.039 & 1.046 \\
  Pilot neuro- & WMD & 1.457 & 1.427 & 1.472 & 1.442 \\
  imaging dataset & CBI & 1.309 & 1.338 & 1.303 & 1.323 \\
  \bottomrule
  \end{tabular}}
  \label{tab8:rmse-short}
\end{table*}

\begin{table*}[h!]
\floatconts
  {tab:rmse-long}
  {\caption{AUROC RMSE for long-range extrapolations.}}
  {\small\begin{tabular}{llS[table-format=2.3]S[table-format=2.3]S[table-format=2.3]S[table-format=2.3]}
  \toprule
  \bfseries Dataset & \bfseries Label & \multicolumn{1}{c}{\bfseries Power law} & \multicolumn{1}{c}{\bfseries GP pow (ours)} & \multicolumn{1}{c}{\bfseries Arctan} & \multicolumn{1}{c}{\bfseries GP arc (ours)}\\
  \midrule
  ChestX-ray14 &  Atelectasis & 1.855 & 1.722 & 2.949 & 2.947 \\
  & Effusion & 3.971 & 3.506 & 3.966 & 3.513 \\
  & Infiltration & 1.616 & 2.079 &  15.350 & 13.470 \\
  Chest X-Ray & Bacterial & 1.067 & 1.089 & 0.208 & 0.210 \\
  & Viral & 3.484 & 3.540 & 2.001 & 1.995 \\
  TMED-2 & PLAX & 0.401 & 0.403 & 1.312 & 1.312 \\
  & PSAX & 0.450 & 0.450 & 0.679 & 0.679 \\
  & A4C & 0.616 & 0.616 & 1.863 & 1.863 \\
  & A2C & 1.974 & 1.971 & 2.285 & 2.286 \\
  \bottomrule
  \end{tabular}}
\end{table*}

\newpage
\section{Results: Curves for Short-Range}
\label{apd:seventh}

\begin{figure*}[htbp!]
\floatconts
  {fig:short-range}
  {\caption{Short-range extrapolations results from ChestX-ray14, Chest X-Ray, BUSI, TMED-2, OASIS-3, and Pilot neuroimaging dataset (top to bottom).}}
  {\includegraphics[width=1\linewidth]{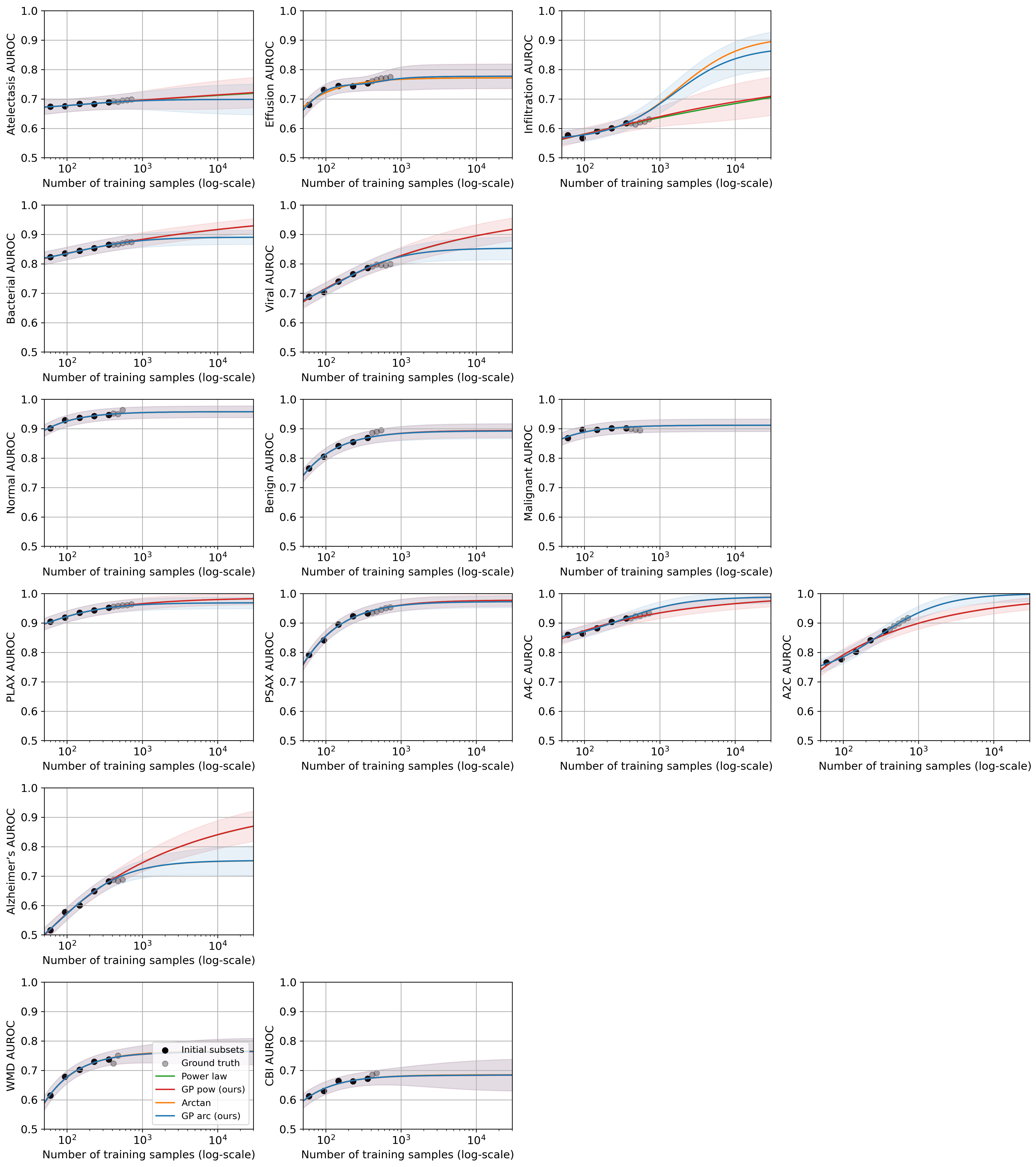}}
\end{figure*}

\newpage
\section{Results: Curves for Long-Range}
\label{apd:eighth}

\begin{figure*}[htbp!]
\floatconts
  {fig:long-range}
  {\caption{Long-range extrapolations results from ChestX-ray14, Chest X-Ray, and TMED-2 (top to bottom).}}
  {\includegraphics[width=1\linewidth]{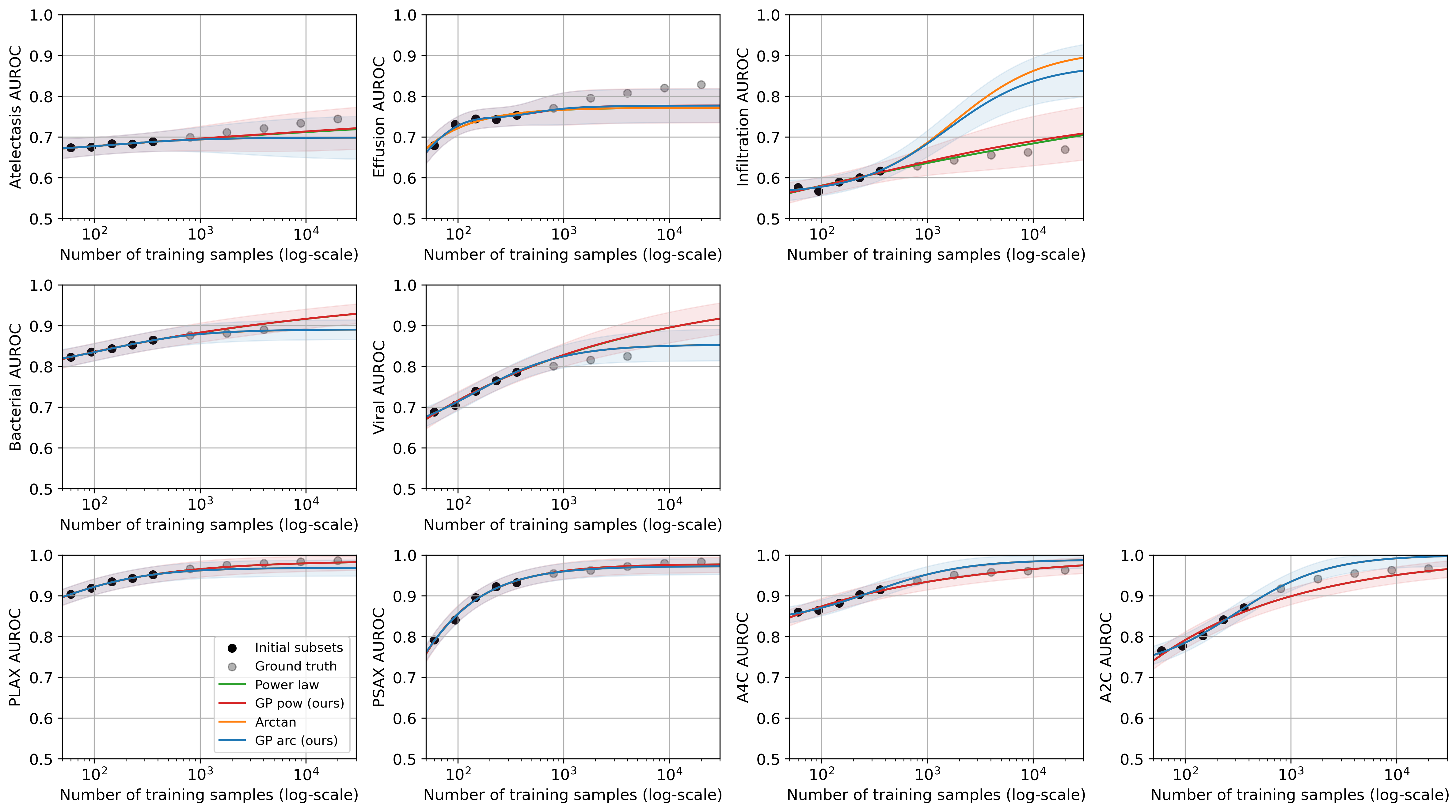}}
\end{figure*}

\newpage
\section{Results: Dense Coverage}
\label{apd:nineth}

\setlength{\tabcolsep}{2pt}
\begin{table*}[htbp!]
\floatconts
  {tab:dense-coverage-80}
  {\caption{Coverage rates for long-range extrapolations, using a target interval of 80\% from our GP model and 100 replicates. Standard deviations generated from 500 bootstrapping rounds to select data-split seeds to average across.}}
  {\footnotesize\begin{tabular}{llr@{.}lr@{.}lr@{.}lr@{.}lr@{.}lr@{.}lr@{.}l}
  \toprule
  & & \multicolumn{4}{c}{5k training samples} & \multicolumn{4}{c}{10k training samples} & \multicolumn{6}{c}{20k training samples} \\
  \bfseries Dataset & \bfseries Label & \multicolumn{2}{c}{\bfseries GP pow} & \multicolumn{2}{c}{\bfseries GP arc} & \multicolumn{2}{c}{\bfseries GP pow} & \multicolumn{2}{c}{\bfseries GP arc} & \multicolumn{2}{c}{\bfseries \makecell{GP pow\\w/o priors}} & \multicolumn{2}{c}{\bfseries GP pow} & \multicolumn{2}{c}{\bfseries GP arc} \\
  \midrule
  ChestX-ray14 & Atelectasis & $100$ & $0\pm0.0\%$ & $97$ & $9\pm1.2\%$ & $100$ & $0\pm0.1\%$ & $77$ & $7\pm2.7\%$ & $12$ & $1\pm2.4\%$ & $100$ & $0\pm0.2\%$ & $29$ & $6\pm2.9\%$ \\
  & Effusion & $67$ & $5\pm2.7\%$ & $61$ & $9\pm3.0\%$ & $9$ & $0\pm2.1\%$ & $6$ & $9\pm1.9\%$ & $0$ & $0\pm0.0\%$ & $0$ & $1\pm0.3\%$ & $0$ & $1\pm0.2\%$ \\
  & Infiltration & $99$ & $4\pm0.8\%$ & $0$ & $0\pm0.0\%$ & $99$ & $1\pm0.9\%$ & $0$ & $0\pm0.0\%$ & $1$ & $0\pm1.0\%$ & $97$ & $8\pm1.3\%$ & $0$ & $0\pm0.0\%$ \\
  TMED-2 & PLAX & $100$ & $0\pm0.1\%$ & $3$ & $9\pm1.6\%$ & $100$ & $0\pm0.0\%$ & $0$ & $1\pm0.2\%$ & $100$ & $0\pm0.0\%$ & $100$ & $0\pm0.0\%$ & $0$ & $0\pm0.0\%$ \\
  & PSAX & $100$ & $0\pm0.0\%$ & $100$ & $0\pm0.2\%$ & $100$ & $0\pm0.0\%$ & $99$ & $2\pm0.8\%$ & $100$ & $0\pm0.2\%$ & $100$ & $0\pm0.2\%$ & $89$ & $5\pm2.4\%$ \\
  & A4C & $99$ & $9\pm0.2\%$ & $0$ & $3\pm0.6\%$ & $99$ & $4\pm0.7\%$ & $0$ & $1\pm0.2\%$ & $94$ & $0\pm1.8\%$ & $94$ & $2\pm1.8\%$ & $0$ & $0\pm0.2\%$ \\
  & A2C & $62$ & $2\pm2.8\%$ & $0$ & $0\pm0.0\%$ & $96$ & $6\pm1.6\%$ & $0$ & $0\pm0.0\%$ & $100$ & $0\pm0.2\%$ & $100$ & $0\pm0.2\%$ & $0$ & $0\pm0.0\%$ \\
  \bottomrule
  \end{tabular}}
\end{table*}
\setlength{\tabcolsep}{6pt}

\section{Results: Single Point Coverage}
\label{apd:tenth}
\begin{table*}[hbtp!]
\floatconts
  {tab:coverage-short}
  {\caption{Single point coverage evaluations for 95\% confidence interval for short-range and long-range extrapolations.}}
  {\small\begin{tabular}{llS[table-format=3.1,table-space-text-post=\%]S[table-format=3.1,table-space-text-post=\%]S[table-format=3.1,table-space-text-post=\%]S[table-format=3.1,table-space-text-post=\%]}
  \toprule
  & & \multicolumn{2}{c}{Short-range extrapolation} & \multicolumn{2}{c}{Long-range extrapolation}\\
  \bfseries Dataset & \bfseries Label & \multicolumn{1}{c}{\bfseries GP pow} & \multicolumn{1}{c}{\bfseries GP arc} & \multicolumn{1}{c}{\bfseries GP pow} & \multicolumn{1}{c}{\bfseries GP arc}\\
  \midrule
  ChestX-ray14 &  Atelectasis & 100.0\% & 100.0\% & 100.0\% & 100.0\% \\
  & Effusion & 100.0\% & 100.0\% & 60.0\% & 60.0\% \\
  & Infiltration & 100.0\% & 100.0\% & 100.0\% & 0.0\% \\
  Chest X-Ray & Bacterial & 100.0\% & 100.0\% & 100.0\% & 100.0\% \\
  & Viral & 100.0\% & 100.0\% &  33.3\% & 100.0\% \\
  BUSI & Normal & 100.0\% & 100.0\% & \multicolumn{1}{c}{---} & \multicolumn{1}{c}{---} \\
  & Benign & 100.0\% & 100.0\% & \multicolumn{1}{c}{---} & \multicolumn{1}{c}{---} \\
  & Malignant & 100.0\% & 100.0\% & \multicolumn{1}{c}{---} & \multicolumn{1}{c}{---} \\
  TMED-2 & PLAX & 100.0\% & 100.0\% & 100.0\% & 100.0\% \\
  & PSAX & 100.0\% & 100.0\% & 100.0\% & 100.0\% \\
  & A4C & 100.0\% & 100.0\% & 100.0\% & 40.0\% \\
  & A2C & 40.0\% & 100.0\% & 60.0\% & 40.0\% \\
  OASIS-3 &  Alzheimer’s & 100.0\% & 100.0\% & \multicolumn{1}{c}{---} & \multicolumn{1}{c}{---} \\
  Pilot neuro- & WMD & 100.0\% & 100.0\% & \multicolumn{1}{c}{---} & \multicolumn{1}{c}{---} \\
  imaging dataset & CBI & 100.0\% & 100.0\% & \multicolumn{1}{c}{---} & \multicolumn{1}{c}{---} \\
  \bottomrule
  \end{tabular}}
\end{table*}

\end{document}